\documentclass[11pt,a4paper]{article}
\usepackage[hyperref]{emnlp-ijcnlp-2019}
\usepackage{times}
\usepackage{latexsym}
\usepackage{graphicx}
\usepackage{url}
\usepackage{enumitem}

\usepackage{url}

\aclfinalcopy % Uncomment this line for the final submission

\title{MIDAS, A Dialog Act Annotation Scheme for Open-Domain Human-Machine Spoken Conversations}

\author{Dian Yu \\
  University of California, Davis \\
  {\tt dianyu@ucdavis.edu} \\\And
  Zhou Yu \\
  University of California, Davis \\
  {\tt joyu@ucdavis.edu} \\}

\date{}

\begin{document}
\maketitle
\begin{abstract}
Dialog act prediction is an essential language comprehension task for both dialog system building and discourse analysis. 
Previous dialog act schemes, such as SWBD-DAMSL, are designed for human-human conversations, in which conversation partners have perfect language understanding ability.
In this paper, we design a dialog act annotation scheme, MIDAS (Machine Interaction Dialog Act Scheme), targeted on open-domain human-machine conversations. MIDAS is designed to assist machines which have limited ability to understand their human partners. MIDAS has a hierarchical structure and supports multi-label annotations. We collected and annotated a large open-domain human-machine spoken conversation dataset (consists of 24K utterances). To show the applicability of the scheme, we leverage transfer learning methods to train a multi-label dialog act prediction model and reach an F1 score of 0.79. %Our experiments demonstrate that the proposed scheme and the dialog act prediction model are suitable for both human annotation and machine prediction.
\end{abstract}

\section{Introduction}
Previous popular dialog act annotation schemes, such as MapTask \cite{maptask}, SWBD-DAMSL \cite{jurafsky}, and ISO \cite{iso1} are designed to understand human-human dialogs. 
Despite the fact that these dialog act schemes are not designed for human-machine conversations, state-of-the-art social conversational systems still use them to train automatic dialog act predictors \cite{gunrock, trento}. 
We believe that an annotation scheme designed specifically for human-machine conversations that addresses their unique features would improve dialog system performance further.

% different distributions
Human-human and human-machine conversations are very different.
Because of the limitation of the machine, humans use different syntax and semantics when talking to a machine than a human. For example, requests such as ``dim the light" are much more frequently seen in human-machine conversations. On the other hand, some labels designed in human-human schemes are not needed in human-machine schemes for machine understanding tasks. For example, separating \textit{Summarize-Reformulate} (e.g. ``Who know what they're doing with that") and \textit{Rhetorical-Questions} (e.g.``Who would steal a newspaper") \cite{jurafsky} in SWBD-DAMSL is not necessary for dialog systems. Moreover, previous schemes annotate conversations on human transcriptions, while in real-time human-machine conversation, transcriptions are not available. Therefore schemes for human-machine dialogs have to operate on unsegmented automatic speech recognition (ASR) outputs. We trained a dialog act predictor model using The Switchboard Dialog Act Corpus (SwDA) annotated with SWBD-DAMSL \cite{jurafsky} and tested on human-spoken dialog system conversations. Even with BERT pre-training, the model's performance is only 47.38\% in prediction accuracy. This low score suggests that only using existing dataset to train models for spoken dialog systems are not applicable. We therefore propose a new annotated human-machine data to solve this problem.

% contributions
In this paper, we propose a hierarchical multi-label dialog act annotation scheme, MIDAS, specifically designed for real-time open-domain human-machine spoken conversations. %We examine the transferablility from human-human to human-machine settings by training on The Switchboard Dialog Act Corpus (SwDA) annotated with SWBD-DAMSL \cite{jurafsky}. 
% In order to improve the performance and reliability of machine prediction, we decide to create an in-domain dataset. 
We annotate real-world human-machine social conversations using the MIDAS scheme. The scheme is easy for human to follow. Two annotators achieve an inter-annotated agreement of $\kappa = 0.94$. 
We train a multi-label dialog act classifier using transfer learning methods and reached a 0.79 in F1 score.
We also share our annotated data and trained models with the research community for the hope of pushing dialog system performance \footnote{\url{https://github.com/DianDYu/MIDAS_dialog_act}}.

\section{Related Work}
\label{sec:related_work}

Previous dialog act annotation schemes are mostly designed for dialogs with a specific task, such as MapTask \cite{maptask} and Verbmobil \cite{verbmobile}. 
There are a few dialog act schemes designed for task-independent conversations, such as the Discourse Annotation and Markup System of Labeling (DAMSL) \cite{damsl} and SWBD-DAMSL \cite{jurafsky}.
SWBD-DAMSL is used to annotate Switchboard \cite{switchboard} corpus, a task-independent telephone conversation corpus.
However, the conversation topics in Switchboard are still limited to 70 pre-defined topics, such as ``air pollution". In this paper, we design a dialog annotation scheme specifically for social chitchat conversations without any topic constraints. 

Most dialog schemes are designed for human-human conversations in the previous research. Because of the recent developments of ASRs and natural language understanding (NLU) technologies, dialog systems, such as Amazon Alexa, have been more and more popular among general users. Therefore, given the discrepancy between human-human and human-machine conversations, 
there is a need to design dialog act schemes for human-machine conversations in order to analyze and support dialog system building.
Khatri \shortcite{cobotda} introduces a human-machine dialog act annotation scheme with 14 tags. However, the scheme is designed for modeling conversation topics instead of training dialog act predictors. The scheme has tags such as \textit{Information Request}, \textit{General Chat}, and \textit{Multiple Goals}, and the annotation is done on unsegmented user utterances. Even though the limited number of tag categories makes annotation more reliable, it may not provide enough information for understanding user intents. For example, tags such as \textit{Multiple Goals} do not provide explicit information on any conversation topic to a dialog manager.
% Furthermore, tags such as \textit{Inappropriate} may be useful for topic modeling but not for understanding current dialog state. 
We propose a dialog act annotation scheme that focuses on improving open-domain dialog system understanding. We also build a dialog act predictor based on the annotated corpus which reached a 0.79 in F1 score. 

% hierarchy; multi-label
Previously, most popular annotation schemes, such as DAMSL, use mutually-exclusive tags \cite{trento} to make annotation process easy and reliable. However, Bunt \shortcite{dit++} argues that conversation utterances are complex. Each functional segment can have four to five functions on average, so dialog act tags should serve multiple functions. Dynamic Interpretation Theory (DIT) \cite{dit} and its extension, DIT++ \cite{dit++} try to solve this problem by supporting multi-dimension and multi-function. The 88 tags are organized in a hierarchically structure and separated into dimension-specific and general-purpose functions. The fifth version of DIT++, ISO \cite{iso1, reiso}, is introduced to incorporate not only linguistic theory but also empirical discourse analysis on real domain-independent conversations. Although much effort has been put in designing ISO, no large dataset annotated using it exists, probably due to the complexity of the scheme and the lack of clear guidelines on how to use contextual information \cite{trento}. Because of the intricacy of open domain social conversations, we propose to build the dialog act scheme to have a hierarchical structure and multi-dimension tags. We also limit the number of dialog acts (23 tags) to make the annotation process feasible (two annotators reached 0.94 in Kapa). We publish the annotated human-machine chatbot corpus that has 24,000 utterances.

\section{MIDAS Annotation Scheme}
\label{sec:scheme}
\begin{figure*}[h]
    \centering
     \includegraphics[width=\textwidth, height=8cm]{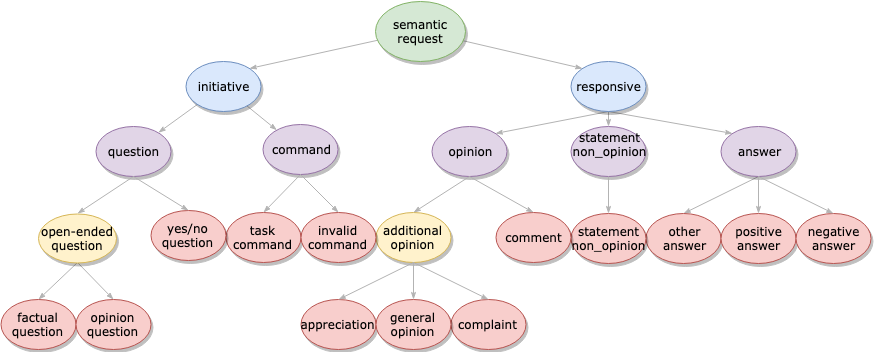}
    \caption{\label{fig:scheme1} Semantic request tree. Scheme types, classes, categories, and sub-categories are in green, blue, purple, and yellow, respectively. Dialog act tags are leaf nodes in red. %Double dashed lines indicate that tags in those categories can overlap (not mutually exclusive). 
    Tags can co-occur in one utterance, except tags under \textit{opinion} and \textit{statement non\_opinion}, \textit{question} and \textit{answer} categories due to semantic and syntactic conflicts. For example, ``User1:Do you watch TV shows? User2: I prefer watching movies." User2 is labeled both \textit{general opinion} and \textit{negative answer}. }
\end{figure*}
We design MIDAS, a contextual hierarchical multi-label dialog act annotation scheme. MIDAS follows DIT++ and ISO \cite{dit++, iso1} to ensure that the scheme is easy to both annotate and train automatic dialog act predictors. 
MIDAS focuses on assisting dialog systems to understand their human users, while previous schemes mainly focus on analyzing human-human dialog. Therefore, besides inheriting labels from previous schemes (such as SWBD-DAMSL and ISO), MIDAS also creates a set of new labels that adapt to the human-machine setting. A complete description of MIDAS is in Appendix \ref{sec:appendix_a}. %MIDAS separates its 23 tags into information request and general 
We discuss the three main features of MIDAS: context completeness, hierarchical structure, and multi-label, respectively as follows. 

\begin{figure*}[h]

    \centering
    \includegraphics[width=\textwidth, height=8cm]{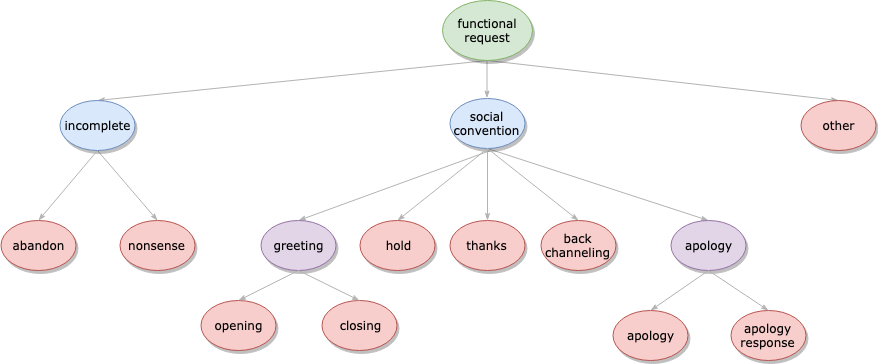}
    \caption{\label{fig:scheme2} Functional request tree. We remove class and category nodes if there is only one label under them.}
\end{figure*}

\subsection{Contextual completeness}
MIDAS relies on contextual information in the annotation process. Specifically, annotators are asked to fill in the ellipsis based on the previous utterances before assigning dialog act tags. For instance, in ``User1: have you read any book recently. User2: the great gatsby", the latter utterance will be completed as ``i read the great gatsby recently". %rather than complete utterance in other context such as ``my favorite boos is the great gatsby". 
The completion process is only for annotators while the original utterances remain unchanged. We leave automatic completion before machine prediction to future work.
%Moreover, ewe emphasize on semantics and rely on the hierarchy. For instance, ``User1: my friend thinks we are living in matrix. User2: she's probably right". User2 is a \textit{general comment}, even though the completed sentence is ``i think she is right", which is \textit{other opinion} given the linguistic form.

Capturing contextual information is more difficult for automated methods compared to human annotators \cite{iso1}. When building dialog act predictor models, we also add the previous user utterances as feature representations. See Section \ref{sec:DA_model} for details.
% For instance, ``the great gatsby" can be completed to ``my favorite book is the great gatsby" (\textit{general\_opinion}) or ``i read the great gatsby recently" (\textit{statement non\_opinion}). 

\subsection{Hierarchical structure}
We design MIDAS to have a tree structure. It has two sub-trees: \textit{semantic request} type (Figure \ref{fig:scheme1}) and \textit{functional request} type (Figure \ref{fig:scheme2}). Under each type, there are classes, categories, and tags arranged in a hierarchical tree structure. Please refer to Figure \ref{fig:scheme1} and \ref{fig:scheme2} for detailed organization. Utterances will be labeled with dialog act tags, which are the leaf nodes. The non-leaf nodes are used to help annotators find the correct dialog act tags.
 %``Request" indicates that the utterance is from a human user interacting with a dialog system. 

\subsubsection{Semantic request} 
Semantic request type captures dialog content, therefore they are essential for dialog topic planning. \textit{Semantic request} separates into \textit{initiative} class and \textit{responsive} class based on whether the user is proposing or continuing a topic. %The selected class indicates the current conversation status.

\noindent\textbf{Initiative class} is especially important in the human-machine setting, because in such unbalanced power setting, the machine has to follow the topic that its human partner proposes. Therefore, understanding whether the user is proposing a new topic with its specific intent is the first step for the system to be coherent. There are two categories, \textit{question} and \textit{command} in the \textit{initiative} class, that are designed to distinguish information request from action request. 

MIDAS first separate \textit{question} into \textit{yes/no question} and \textit{open-ended question} based on syntax. Such separation helps the system to generate coherent response. For example, system responses are more natural to start with words, such as ``yes" or ``no" when replying to \textit{yes/no question}. Then MIDAS further separates \textit{open-ended question} into \textit{factual question} and \textit{opinion question} based on different types of information that users seek. The system need to search different knowledge bases based on the tag. For example, \textit{factual questions} requires factual information from knowledge graphs such as %Evi \footnote{\url{https://www.evi.com/}} and 
Wikipedia%\footnote{\url{https://www.wikipedia.org/}}
, while \textit{opinion question} requires information from corpora with opinionated material such as Twitter database.  

Different from \textit{question}, \textit{command} conveys orders and is particularly popular in human-machine dialogs.
The system needs to follow users' command, whether implicit or explicit, because the system has less power in the conversation with human. Therefore, unlike ISO and other schemes, MIDAS combines command types, such as \textit{direct request}, \textit{indirect request}, and \textit{suggestion}. Combining these tags also simplifies the annotation scheme. In addition, we add an extra tag under \textit{command}, 
\textit{invalid command}, which tailors to smart devices. Users sometimes produce commands that are out of the system's capability. For example, users may want to control the device hardware that the dialog system does not have access to currently. The system will want to identify these utterances and handle them separately.

\noindent\textbf{Responsive class} indicates that the utterance is a continuation of the previous topic. SWBD-DAMSL notices that opinions are often followed by other opinions, whereas statements are followed by backchannels \cite{jurafsky}. This distinction may not be beneficial to human-human conversations \cite{jurafsky} as humans do not need to explicitly distinguish between the two tags to generate corresponding responses.
However, knowing whether an utterance is a statement or an opinion is essential for a system to generate appropriate responses. In addition, a quick and definite reply to the previous utterance proposal or question can benefit dialog planning.
% The difference between \textit{answer} and the other two categories is that \textit{answer} explicitly targets a question form, indicating a quick reply to the previous utterance. This category is beneficial for dialog planning given the current proposal or a question. 
%In addition, answers responding to a question or a proposal explicitly indicate the preference of the utterance. 
Hence, MIDAS further breaks the responsive class into \textit{opinion}, \textit{statement non-opinion}, and \textit{answer}, based on conversation history.

MIDAS separates the \textit{opinion} category into the \textit{additional opinion} subcategory and the \textit{comment} tag because we observed examples such as ``User1: my friend thinks we are living in matrix. User2: she's probably right". 
User2 comments on the previous utterance without contributing extra information. \textit{Comment} often indicates an utterance of simply reply, without explicit feedbacks.
%\textit{Comment} indicates that the user acknowledges and builds the utterance on the previous system utterance while \textit{general opinion} does not. 
MIDAS separates three types of opinions: \textit{appreciation}, \textit{complaint}, and \textit{general opinion}, because having the sentiment valence of the opinion can help the system plan the dialog better. For dialog systems, understanding whether the user is complaining or praising them is essential information to plan for dialog policies. %Appendix \ref{sec:appendix_a} describes these tags in detail along with examples.

We also break down \textit{answer} into \textit{positive answer}, \textit{negative answer}, and \textit{other answer}, based on utterance sentiment. One caveat is that utterances, such as ``why not", contain negative words but are actually positive answer for questions such as ``can we talk about movies". Such phenomena suggests that automatic dialog act prediction models require semantic understanding and need to incorporate context in feature representation. %These utterances rely on semantic contextual information to obtain accurate annotation. %We define \textit{positive\_answer}, \textit{negative\_answer}, and \textit{other\_answer} to provide extra information together with other co-occurring labels to assist understanding an utterance for dialog management.

\subsubsection{Functional request}
\textit{Functional request} type %provides discourse level information to help 
helps dialog systems achieve discourse level coherence.
%Compared to \textit{information request}, general request focus on discourse level instead of task level intents.
We define \textit{incomplete}, \textit{social convention}, and \textit{other} classes under the \textit{functional request} type. %These types are essential to the conversational status, but do not add more semantic information to the current conversation compared to labels in the Information request type.

\noindent\textbf{Incomplete class} describes utterances that are not complete. There are two types of \textit{incomplete}, \textit{abandon} and \textit{nonsense}. In real-world settings, human users can be cut off due to issues such as background noises and long pauses. These cases are labeled as \textit{abandon}.  
In comparison, \textit{nonsense} is used to label utterances that human annotators cannot understand. These utterances usually have many ASR errors.  %Instead of grammatical errors from human users (which can be captured by a language model) or errors from dropping words or syllables (e.g. ``scure stories" instead of ``obscure stories"), both of which can still be inferred given the context, this label focuses on the semantic meaning and 
The system can understand both \textit{abandon} and \textit{nonsense} utterances better by asking users to repeat. However, MIDAS still separates %\textit{abandon} and \textit{nonsense}, 
the two, because if the utterance has an \textit{abandon} tag, such as ``i think", the system can give users more specific instructions such as ``take your time". Such instructions are not applicable for  \textit{nonsense} utterances.

\noindent\textbf{Social convention class}
is similar to the \textit{social obligations management} and \textit{discourse structure management} dimensions in ISO \cite{iso1}. There are \textit{opening}, \textit{closing}, \textit{thanks}, \textit{apology}, \textit{apology response}, \textit{hold}, and \textit{back channeling} to provide discourse level information.

% Different from human-human conversations, utterances such as ``i think i'm done with it" and ``that's it" signals closing a conversation besides ``good night" and ``bye" in human-human conversations. Dialog context is critical in understanding these utterances. 

Finally, utterances that cannot be assigned to any other tag in this hierarchical structure are labeled as an \textit{other} tag. 

\subsection{Multi-label support}
\label{sec:multi}
Compared to single-label schemes, multi-label schemes capture different dimensions and functions, which support dialog system building and discourse analysis better. For example,\\ \par

User1: \textit{what books have you read recently}\par
User2A: \textit{i haven't read any}\par
User2B: \textit{i don't want to talk about books}\par
User2C: \textit{i prefer watching movies}\\

Users may use different sentences to express a \textit{negative answer} intent. If the annotation scheme is single-label, the above three sentences cannot be differentiated. Having an extra label to capture other semantic information besides \textit{negative  answer} will benefit dialog system building. For instance, User2B has the additional \textit{task command} intent to end the current topic compared to User2A. User2C has the \textit{general opinion} intent to initiate a different topic. While the dialog system may not need to change the topic if the utterance is User2A, but will have to change if it is User2B.%, and will change to a specific topic, movies, if it is User2C.

%In contrast, only predicting an \textit{general\_opinion} intent does not provide semantic information on whether the user is having an opinion towards the current topic (e.g. ``I find the black swan very interesting to read") or towards a different topic. On the other hand, having exactly one label for each utterance requires detailed rules to assign priority to labels if multiple ones could apply. Such a requirement may add ambiguity to both human annotators and machine prediction models. 

%Besides more complete semantic understanding, the multi-label scheme also considers linguistic forms to give hints to the format of the generated response. For instance, all the utterances ``let's talk about star wars" (\textit{statement\_non-opinion}), ``how do you like star wars" (\textit{opinion\_question}), and ``what do you know about star wars" (\textit{factual\_question}) are suggesting that the user wants to talk about a specific topic (star wars). It is important for the dialog manager to understand that this is a \textit{command} intent. More importantly, the NLG modules need to understand that interactor would actually expect responses that are initiating a conversation, a personal review, and some factual information, respectively from the three different example utterances. Response texts can be generated accordingly with targeted knowledge bases. Therefore, multi-label support is desirable to construct a natural and coherent conversation. 

SWBD-DAMSL allows a utterance to be tagged as \textit{double labels} and lists the preferred tag first \cite{jurafsky}. However, the rules designed to order these tags rely on heuristics and are not explicit to follow.
Our proposed scheme, MIDAS, also allows multiple tags with a clear priority but do not require annotators to order the tags. The scheme is thus easier to follow and more reliable. %As we believe, in different context, different dialog acts is more important for understanding, but ask individual human to make the decision is difficult.
In MIDAS, except for two exclusive category pairs (\textit{opinion} and \textit{statement non-opinion}, \textit{question} and \textit{answer}), labels in each category can co-occur with another. However, we restrict the maximum number of tags for each utterance to be two, in order to reduce the complexity in the annotation process, as well as machine prediction. %For simplicity, we do not consider complex phenomenon, such as sarcasm. 
%The most frequently co-occurred labels in the annotated data are listed in Appendix \ref{appendix:multi} along with the number of occurrences.

If there are more than two dialog act tags applicable to an utterance, we choose the two that are most useful for dialog planning (without ordering). Due to the unbalanced power of human-machine conversations, MIDAS prefers dialog act categories in the following order: \textit{answer}, \textit{command}, \textit{opinion}, \textit{statement non\_opinion}, and \textit{question}.
For example, ``User1: what do you want to talk about? User2: how about the financial market". User2's utterance can be tagged as \textit{task command}, \textit{opinion question}, and \textit{other opinion}. Among the three tags, \textit{task command} and \textit{opinion question} are more important for the system to direct the conversation towards a specific topic so we only keep these two.

\section{Dataset and Annotation Process}
We collected 380K human-machine conversations using Gunrock, the 2018 Alexa Prize winning social bot \cite{gunrock}.
Table \ref{table:chat_example} shows an sample conversation. Two annotators reached an inter-annotator agreement of $\kappa = 0.94$ on 1,185 segmented utterances. Then they annotated the rest of the data separately. In total, they annotated 468 conversations, including 24K segmented sentences on both user and system, among which 12.9K segments are from users. \textit{general opinion} and \textit{statement non\_opinion} are the most frequent tags. For multiple labels, (\textit{positive\_answer}, \textit{command}) and (\textit{negative\_answer}, \textit{command}) are the most frequent co-occur tags. Detailed annotation results including dialog act tags distribution are shown in Appendix \ref{sec:appendix_a} and Appendix \ref{appendix:multi}.
\begin{table}[h!]
\small
\def\arraystretch{1}
\begin{tabular}{l p{0.75\linewidth}}
%\hline
\textbf{USER1}: & what do you want to talk about \\
\textbf{USER2}: & what can you tell me what the top books are right now\\
\textbf{USER1}: & i am so excited to talk to you about books. i'm actually a pretty big bookworm, and i love to read when i'm not chatting\\
\textbf{USER2}: & oh [SEG] what are some titles of the books you've read\\
\textbf{USER1}: & recently, i'm reading the great gastby. it's really thought provoking, and i can see why some people call it the great american novel. how about you? what book do you like? \\
% \label{sample:seg} 
\textbf{USER2}: & i have \{not\} read a book in a while [SEG] do you have recommendations in the sci fi  \\
%\hline
\end{tabular}
\caption{An example conversation between a machine (USER1) and a human (USER2). The word ``not" is dropped in the last sentence due to ASR errors.} %User utterances (USR) are segmented by the segmentation model and corrected by annotators during annotation. The segmentation token is represented as [SEG]. }
\label{table:chat_example}
\end{table}

For dialog act prediction, user utterances in human-machine dialogs are ASR outputs and have no punctuation. Therefore, we train a model to segment utterances into complete semantic units for pre-process. We then perform dialog act prediction on each segmented unit \cite{stolcke}. 
% User utterances in human-machine dialogs are ASR outputs and have no punctuation. %So different from annotated segmentations \cite{stolcke}, 
% Therefore, we first train a model to segment these utterances into complete semantic units and then perform dialog act prediction on each segmented unit. 
Previous research detects sentence boundaries by predicting the exact punctuation in the training dataset \cite{nncrf_punc}. %Some model also considers audio information to improve detection performance\cite{s2s_seg}. 
However, correct punctuation also relies on deep semantic understanding beyond the sentence surface forms. A misused question mark can lead the dialog act model to predict a sentence as a question. So following \citet{global_features}, we only predict the boundary of the sentence instead of predicting punctuation to avoid introducing errors.

Because it is expensive to annotate sentence boundaries, we use the Cornell Movie-Quotes Corpus \cite{cornell_movie_data} to train a sentence segmentation model.
The Cornell dataset contains 300K utterances from movie transcripts. We reformat the transcripts by replacing punctuation to sentence breaker tokens (denotes as [SEG]). %We only preserves periods in the data to denote sentence boundaries. %We replaced all the punctuation with a special [SEP] token and lower-cased all the words for the target corpus. [SEP] is removed in the parallel source corpus. The goal of the segmentation model is to predict the [SEP] token in the correct position in a complex sentence. 
We then trained a sequence-to-sequence (seq2seq) model to predict sentence breaker tokens similarly to \citet{s2s_seg} and  \citet{nmt_punct}. Both the encoder and the decoder are 2-layer 500-dimension bi-LSTM. In addition, the decoder uses global attention and input feed \cite{attention} with beam search. The input of the model is a reformatted sentence, and the output is the same sentence with added sentence breaker tokens. An example can be seen in the last \textit{USER2} utterance in Table \ref{table:chat_example}. Word embeddings are pre-trained with fastText \cite{fasttext} using Common Crawl.  %The model is optimized with stochastic gradient descent (SGD).
% In vanilla beam search, even in the training the input output are exact same words, during testing, the output may still slightly deviate from the input.  So we also post-process to ensure the output is identical to the input except for the inserted predicted break tokens. 
We evaluate the segmentation model on human labeled 2K human utterances of collected data. The segmentation model achieves $84.43\%$ in micro F1 score, $84.97\%$ in precision, and $84.57\%$ in recall. We apply the trained segmentation model on the entire collected dataset to obtain segmented sentences. All the dialog act annotation and predictions are done on the automatic segmentation results. 
% Note that during annotation, we also let the annotators to correct segmentation errors. 

\section{Dialog Act Prediction}
\label{sec:DA_model}
We formulate the dialog act prediction problem as a multi-label classification problem. Building on previous work on text classification, we use an encoder-decoder model with two major modifications. We incorporate context in feature design and have one or two labels as output. In addition, we leverage both unlabeled data and annotated data in the transfer learning process.

\subsection{Baseline model}
RNN models have shown promising results on text classification \cite{lstmcnn}. Our baseline model uses a 2-layer Bi-LSTM to encode the context representation and a multi-layer perceptron (MLP) to decode the output. For multi-label prediction, we use a binary cross-entropy objective function. During testing, we choose the labels with the highest two values predicted from the MLP as the potential output and filter them with an empirical threshold (0.5) to decide to keep both labels or just the one with the highest probability. 

\subsection{Context representation}
\label{preprocess_data}
Contextual information plays an important role in dialog act prediction \cite{fb, cobotda}. We consider two methods to represent previous turns: the actual utterance (text), and the dialog act of the utterance (DA).
For each method, the most recent segmented sentence unit from each speaking party is considered as the history. We append the last segmented system unit (sys\_unit), the previous segmented user unit (user\_prev), and the current segmented user unit (user\_cur) as $sys\_unit$ $<$$u\_p$$>$ $user\_prev$ $<$$u\_c$$>$ $user\_cur$ where $<$$u\_p$$>$ and $<$$u\_c$$>$ are special tokens to separate utterances. For instance, to predict the dialog act for the segment ``do you have recommendations in the sci fi" in the last \textit{USER2} utterance in Table \ref{table:chat_example},
% for ``maybe lebron james" as in the conversation\\
%   A: \textit{who do you think is the best basketball player}\par
%   B: \textit{i don't know [SEG] maybe lebron james}\\
the context representation is formed as \textit{what book do you like $<$u\_p$>$ i haven't read a book in a while $<$$u\_c$$>$ do you have recommendations in the sci fi}. However, if the current utterance is the first one in the current turn, i.e. there is no contextual information, we use an empty token for $usr\_prev$ instead.

Another method to incorporate history is to replace the actual previous segment unit with its dialog act labels (if there are two labels for one segment, we combine both labels). The results for these two methods are shown in Table \ref{result:basic}.

\subsection{Transfer learning}
\label{transfer_learning}
We experimented with two methods to leverage more data. One is an unsupervised task on domain adaption and the other is a supervised dialog act predictor trained on SwDA \cite{jurafsky}. %Recent advances such as fastText and BERT \cite{bert} suggested the advantages of using dynamic pre-trained word embeddings on natural language processing tasks. 
%For the base model (the LSTM model), we map word embeddings to fastText and train the embeddings with a Bi-LSTM and a MLP jointly. 

For domain adaption, we started with the BERT based model trained on Wikipedia \cite{bert} % \footnote{\url{https://github.com/google-research/bert}} 
to leverage contextual word embeddings from a large language model. %We reformatted the context representation by concatenating the context and the current segmented unit. 
However, one potential drawback of using BERT pre-trained on textual data is its domain difference from conversational data. Inspired by Siddhant et al. \shortcite{transfer}, we use 50 million unlabeled segmented utterances collected from 380K conversations from Gunrock to fine-tune the BERT language model before training on the classification task. 

In addition to pre-trained word-embeddings from language models, we leverage annotated datasets. We automatically map 42 tags from SWBD-DMSL to our 23 tags. The detailed mapping can be found in Appendix \ref{mapping}. We remove all the punctuation (except apostrophes) and non-verbal information such as ``$<$laugh$>$" from the carefully annotated dataset. We also drop sentences with dialog act that is not applicable to ours such as \textit{3rd-party-talk}.  %We reformatted the data so that the utterances are in turns, similar to our collected data. 
Because there are only 4 utterances labeled with two tags out of 386K original utterances in SwDA, we consider this as a single-label dataset. After pre-processing, we extract a total of 200K annotated utterances using context representation explained in Section \ref{preprocess_data}. We train a single-label prediction model based on BERT before fine-tuning it on multi-label prediction with our annotated data.

\section{Experiments}
\label{sec:experiments}
\textbf{Setting} Unlike human-human conversations, in which the interlocutors share common patterns, machine and user utterances in a dialog system are very different from each other. For instance, machine utterances may have punctuation, contain no ASR errors, and have limited vocabulary and syntax compared to user utterances. Due to the differences between user and machine utterances in human-machine conversations, we cannot combine them during prediction. The main purpose of having a dialog act predictor is for dialog system to understand user intent better. Therefore, we build a dialog act prediction model on user utterances only.
After pre-processing (refer to Section \ref{preprocess_data} for details), there are 12.9K user segments. $13.78\%$ of them have two labels. We use 10.3K for training and 2.6K for testing. The rest 11.1K annotated machine segments are used as context.

\noindent\textbf{Models.} We implemented 11 models as follows:

% \begin{itemize}[noitemsep,nolistsep]
%   \item \textbf{LSTM} means model trained by LSTMs. LSTM-text uses text and LSTM-DA uses dialog act as the context.
%   \item \textbf{BERT} stands for transformer models with pre-trained BERT language model. BERT-text and BERT-DA uses text and dialog act as the context, respectively. BERT-no\_context means the model does not consider the context, i.e. use the current segment only. BERT-DA+text uses both text and dialog act as the context.
%   \item \textbf{BERT\_F} means transformer models wtih pre-trained BERT language model fine-tuned on unsupervised in-domain data. We consider text, dialog act, and both text and dialog act as the context for BERT\_F-text, BERT\_F-DA, and BERT\_DA+text, respectively.
%   \item \textbf{BERT-SwDA} means transformer models with pre-trained BERT language model fine-tuned on SwDA task using text as context. BERT-SwDA\_F is further fine-tuned based on BERT\_F and BERT-SwDA is further fine-tuned based on BERT.

% \end{itemize}
We use \textbf{LSTM} to represent the baseline model trained with LSTMs. We use \textbf{BERT} to represent transformer models with a pre-trained BERT language model. According to different transfer learning methods described in Section \ref{transfer_learning}, \textbf{BERT\_F} is a pre-trained BERT language model fine-tuned on unlabeled in-domain data, whereas \textbf{BERT\_SwDA} is a pre-trained BERT language model fine-tuned on labeled SwDA task. Combining these two methods, \textbf{BERT-SwDA\_F} fine-tunes on both the unlabeled and labeled tasks. After fine-tuning, the models are trained on our annotated data with MIDAS scheme. To evaluate the impact of context representation for the above models, we use \textbf{-text} and \textbf{-DA} to represent using text and dialog act as the context, respectively. In addition, we denote \textbf{-no\_context} to predict on the current utterance only without using any context.

% For LSTM, BERT base (BERT), and BERT fine-tuned on unsupervised in-domain data (BERT\_F), we experimented with using text (-text) and dialog act (-DA) as the context. We also considered both dialog act and text (-DA+text) to train on BERT and BERT\_F. For pre-training on a similar classification task, we trained models with SwDA with BERT model (BERT-SwDA) and BERT\_F model (BERT-SwDA\_F) before fine-tuning on in-domain prediction tasks.

% We experimented with using both text and dialog act as the context

% \textit{-text} refers to using utterance as the context and \textit{-DA} refers to using dialog act as the context. \textit{BERT\_F} is a language model fine-tuned on \textit{BERT} (BERT base language model) with unsupervised human-machine conversation. \texit{SwDA\_F} uses pre-trainined classification model on SwDA trained with \textit{BERT\_F}. In comparison, \textit{SwDA\_B} is trained on SWBD with \textit{BERT}. \textit{DA+text} refers to considering both text and dialog act lables as input features.

\noindent\textbf{Implementation details.} The baseline dialog act prediction model uses a 2-layer Bi-LSTM with a hidden size of 500. The LSTM layers use a dropout rate of 0.3. We optimize the model with Adam optimizer \cite{adam}.
For the transformer models, we use 12 layers with 12 attention heads and a hidden size of 768. All the fully connected layers use a dropout rate of 0.1.

Because one data sample may have two labels in our annotation, we calculate precision, recall, and F1 on each sample and then average them across all samples (micro F1). 

\section{Results and Analysis}
  \begin{table} 
    \begin{center} 
     \begin{tabular}{|l | c | c | c|}
     \hline
      & Pre(\%)& Rec(\%) & F1(\%) \\ [0.5ex] 
     \hline
     LSTM-text & $75.94$ & $75.91$ & $75.51$ \\ 
     \hline
     LSTM-DA & $75.83$ & $73.48$ & $73.77$ \\ 
     \hline
     BERT-text & $79.57$ & $79.31$ & $79.11$ \\
     \hline
     BERT-DA & $79.29$ & $76.12$ & $76.87$\\ 
     \hline
     BERT-no\_context & $73.88$ & $70.43$ & $71.30$\\ 
     \hline
     BERT-DA+text & $79.79$ & $79.47$ & $79.28$ \\ 
     \hline
     BERT\_F-text & $79.83$ & $79.64$ & $79.40$ \\ 
     \hline
     BERT\_F-DA & $79.30$ & $76.15$ & $76.89$ \\ 
     \hline
     BERT\_F-DA+text & $\textbf{79.93}$ & $79.61$ & $\textbf{79.44}$ \\ 
     \hline
     BERT-SwDA & $79.26$ & $76.43$ & $78.98$ \\ 
     \hline
      BERT-SwDA\_F & $79.58$ & $\textbf{79.76}$ & $79.28$ \\ 
     \hline
    %  DA+text\_F & $\textbf{79.93}$ & $79.61$ & $\textbf{79.44}$ \\ 
    %  \hline
    %  combine\_B & $79.79$ & $79.47$ & $79.28$ \\ 
    %  \hline
    \end{tabular}
    \caption { BERT\_F-DA+text achieves the best precision and F1 score. Results reported are an averaged score of six different random seed runs.} \label{result:basic}
    \end{center}
  \end{table}
  
\label{sec:results}
Table~\ref{result:basic} describes the experimental results on all 11 models. Transformer models using BERT embeddings (BERT-text) outperform Bi-LSTM models with pre-trained word embeddings (LSTM-text) by a large margin (from 75.51\% to 79.11\% in F1). If we further fine tune the BERT language model on an unsupervised training task with similar data distribution (BERT\_F-text), the classification result further improves from 79.11\% to 79.40\% in F1. This is consistent with previous research on in-domain pre-training \cite{transfer}. However, the performance improvement is not statistically significant. One possible reason is that models pre-trained on a very large text dataset, such as Wikipedia, already encodes sufficient semantics for dialog act prediction. Therefore, fine-tuning the model on a more domain aligned data set does not improve the performance drastically.

We found that incorporating context improves the model performance. Adding text information as context improves the BERT model from 71.30\% to 79.11\% in F1. We also compare the impact of different context embedding methods on dialog act classification performance. 
The results show that replacing text with dialog act achieves a high precision, but suffers from a low recall. 
This is because an utterance can have multiple intents while dialog act itself does not provide enough context information to achieve accurate prediction. 
For example, when ``i don't think so" is a response to a simple \textit{yes/no question} such as ``have you read the book", it is a \textit{negative\_answer}. But if it is a response to a more complex \textit{yes/no question}, such as ``do you want to talk about books", then it has two tags, \textit{command} and \textit{negative\_answer}. The latter conveys user's implicit request on changing the topic. Therefore, only using dialog act as context could lead to high recall but low F1.
We found combining both previous segment's dialog act label 
and its surface text together achieves the best performance in F1 (79.44\%). However the performance improvement over including text only is not statistically significant. This suggests that dialog act and text may have more overlapped information than complimentary information. 

We also found that fine-tuning the model using the supervised dialog act prediction task on the SwDA data did not improve performance in F1 but improved recall slightly. The reduced performance may be due to the data difference. Even though both datasets are open domain conversational data, SwDA task uses pre-processed Switchboard data that does not have ASR errors. Moreover, SwDA is human-human conversations, and they are more coherent and consistent compared to human-machine conversations. Another reason is that SwDA dataset has exactly one label for each utterance. When fine-tuning on our multi-label task, the pre-trained single-label model may tend to predict more labels to quickly reduce loss but fail to learn better representations.

We further looked into the errors from the best model (\textit{BERT\_F-DA+text}) and found that the model confuses \textit{statement non-opinion} and \textit{general opinion}. This is most likely caused by only including one turn context. Sometimes, users would have questions that breaks the conversation flow, such as ``can you say it again clearly". The model needs to consider not only this utterance but also the turns before that to perform dialog act prediction. We plan to incorporate longer context in future work. In addition, some of the \textit{nonsense} sentences are misclassified as \textit{statement non\_opinion} such as ``it doesn't outside break a car". It is also worth noting that some incorrectly segmented units resulted in inaccurate dialog act prediction.
% always predicts \textit{general\_opinion}. This is due to the fact that the model does not consider context in the previous turns when the system asks the actual question such as ``what book have you read recently". 

\section{Conclusion and Future Work}
\label{sec:conclusion}
We propose a dialog act scheme designed for open-domain human-machine conversational systems, MIDAS. MIDAS is a hierarchical annotation scheme that supports multiple labels. %It specifically focuses on separating out commands to cater human-machine conversations. 
We annotated 24K sentences from a human-machine social conversation data using MIDAS. We also trained dialog act classification models based on the annotated dataset. We tested different transfer learning techniques to improve model performance. We found that fine-tuning using the pre-trained BERT embedding plus the unannotated target human-machine conversation improved model performance. But fine-turning the model on a supervised dialog act task with human-human data did not improve model performance. 

In the future, we plan to combine dialog act and parsing in a multi-task learning setting, so the dialog act model can borrow information from syntatic and semantic parsing representations. In addition, we would like to test if training the utterance segmentation and dialog act prediction together can improve model performance.

% \section*{Acknowledgments}

% The acknowledgments should go immediately before the references.  Do
% not number the acknowledgments section. Do not include this section
% when submitting your paper for review. \\

% \noindent {\bf Preparing References:} \\

% Include your own bib file like this:
% {\small\verb|\bibliographystyle{acl_natbib}|
% \verb|\bibliography{emnlp-ijcnlp-2019}|}

% Where \verb|emnlp-ijcnlp-2019| corresponds to the {\tt emnlp-ijcnlp-2019.bib} file.
\bibliography{emnlp-ijcnlp-2019}

\begin{thebibliography}{25}
\expandafter\ifx\csname natexlab\endcsname\relax\def\natexlab#1{#1}\fi

\bibitem[{Alexandersson et~al.(1998)Alexandersson, Buschbeck-Wolf, Fujinami,
  Kipp, Koch, Maier, Reithinger, Schmitz, and Siegel}]{verbmobile}
Jan Alexandersson, Bianka Buschbeck-Wolf, Tsutomu Fujinami, Michael Kipp,
  Stephan Koch, Elisabeth Maier, Norbert Reithinger, Birte Schmitz, and Melanie
  Siegel. 1998.
\newblock Dialogue acts in verbmobil-2 - second edition.
\newblock In \emph{DFKI Saarbrücken}.

\bibitem[{Bunt(1997)}]{dit}
Harry Bunt. 1997.
\newblock Dynamic interpretation and dialogue theory.

\bibitem[{Bunt(2009)}]{dit++}
Harry Bunt. 2009.
\newblock The dit++ taxonomy for functional dialogue markup.
\newblock In \emph{AAMAS 2009 Workshop, Towards a Standard Markup Language for
  Embodied Dialogue Acts}, pages 13--24.

\bibitem[{Bunt et~al.(2010)Bunt, Alex, Carletta, woong Choe, Fang, Hasida, Lee,
  Petukhova, Popescu-belis, Romary, Soria, and Traum}]{iso1}
Harry Bunt, Jan Alex, Jean Carletta, Jae woong Choe, Alex~Chengyu Fang, Koiti
  Hasida, Kiyong Lee, Volha Petukhova, Andrei Popescu-belis, Laurent Romary,
  Claudia Soria, and David Traum. 2010.
\newblock Towards an iso standard for dialogue act annotation.
\newblock In \emph{1st Proceedings of Alexa Prize}.

\bibitem[{Bunt et~al.(2017)Bunt, Petukhova, and Fang}]{reiso}
Harry Bunt, Volha Petukhova, and Alex Fang. 2017.
\newblock Revisiting the iso standard for dialogue act annotation.
\newblock In \emph{Proceedings 13th Joint ISO - ACL Workshop on Interoperable
  Semantic Annotation (isa-11)}, pages 37--50.
\newblock Research Unit(s) information for this publication is provided by the
  author(s) concerned.

\bibitem[{Chen et~al.(2018)Chen, Yu, Wen, Yang, Zhang, Zhou, Jesse, Chau,
  Bhowmick, Iyer, Sreenivasulu, Cheng, Bhandare, and Yu}]{gunrock}
Chun-Yen Chen, Dian Yu, Weiming Wen, Yi~Mang Yang, Jiaping Zhang, Mingyang
  Zhou, Kevin Jesse, Austin Chau, Antara Bhowmick, Shreenath Iyer, Giritheja
  Sreenivasulu, Runxiang Cheng, Ashwin Bhandare, and Zhou Yu. 2018.
\newblock Gunrock: Building a human-like social bot by leveraging large scale
  real user data.
\newblock In \emph{2nd Proceedings of Alexa Prize}.

\bibitem[{Cho et~al.(2015)Cho, Kilgour, Niehues, and Waibel}]{nncrf_punc}
Eunah Cho, Kevin Kilgour, Jan Niehues, and Alexander~H. Waibel. 2015.
\newblock Combination of nn and crf models for joint detection of punctuation
  and disfluencies.
\newblock In \emph{INTERSPEECH}.

\bibitem[{Core and Allen(1997)}]{damsl}
Mark~G. Core and James~F. Allen. 1997.
\newblock Coding dialogs with the damsl annotation scheme.
\newblock In \emph{Working Notes of the AAAI Fall Symposium on Communicative
  Action in Humans and Machines}, pages 28--35.

\bibitem[{Danescu-Niculescu-Mizil and Lee(2011)}]{cornell_movie_data}
Cristian Danescu-Niculescu-Mizil and Lillian Lee. 2011.
\newblock Chameleons in imagined conversations: A new approach to understanding
  coordination of linguistic style in dialogs.
\newblock In \emph{Proceedings of the Workshop on Cognitive Modeling and
  Computational Linguistics, ACL 2011}.

\bibitem[{Devlin et~al.(2018)Devlin, Chan, Lee, and Toutanova}]{bert}
Jacob Devlin, Ming-Wei Chan, Kenton Lee, and Kristina Toutanova. 2018.
\newblock Bert: Pre-training of deep bidirectional transformers for language
  understanding.
\newblock \emph{arXiv preprint arXiv:1810.04805, 2018.}

\bibitem[{Favre et~al.(2008)Favre, Hakkani-Tur, Petrov, and
  Klein}]{global_features}
B.~Favre, D.~Hakkani-Tur, S.~Petrov, and D.~Klein. 2008.
\newblock Efficient sentence segmentation using syntactic features.
\newblock In \emph{2008 IEEE Spoken Language Technology Workshop}, pages
  77--80.

\bibitem[{Godfrey et~al.(1992)Godfrey, Holliman, and McDaniel}]{switchboard}
John~J. Godfrey, Edward~C. Holliman, and Jane McDaniel. 1992.
\newblock Switchboard: telephone speech corpus for research and development.
\newblock In \emph{[Proceedings] ICASSP-92: 1992 IEEE International Conference
  on Acoustics, Speech, and Signal Processing}, volume~1, pages 517--520 vol.1.

\bibitem[{Jurafsky et~al.(1997)Jurafsky, Shriberg, and Biasca}]{jurafsky}
Dan Jurafsky, Liz Shriberg, and Debra Biasca. 1997.
\newblock {Switchboard SWBD-DAMSL shallow-discourse-function annotation coders
  manual}.
\newblock Technical Report Draft 13, University of Colorado, Institute of
  Cognitive Science.

\bibitem[{Khatri et~al.(2018)Khatri, Goel, Hedayatni, Metanillou, Venkatesh,
  Gabriel, and Mandal}]{cobotda}
Chandra Khatri, Rahul Goel, Behnam Hedayatni, Angeliki Metanillou, Anushree
  Venkatesh, Raefer Gabriel, and Arindam Mandal. 2018.
\newblock Contextual topic modeling for dialog systems.
\newblock In \emph{IEEE 2018 Spoken Language Technology}.

\bibitem[{Kingma and Ba(2014)}]{adam}
Diederik~P. Kingma and Jimmy Ba. 2014.
\newblock \href {http://arxiv.org/abs/1412.6980} {Adam: {A} method for
  stochastic optimization}.
\newblock \emph{CoRR}, abs/1412.6980.

\bibitem[{Klejch et~al.(2017)Klejch, Bell, and Renals}]{s2s_seg}
O.~Klejch, P.~Bell, and S.~Renals. 2017.
\newblock Sequence-to-sequence models for punctuated transcription combining
  lexical and acoustic features.
\newblock In \emph{2017 IEEE International Conference on Acoustics, Speech and
  Signal Processing (ICASSP)}, pages 5700--5704.

\bibitem[{Liu et~al.(2017)Liu, Han, Tan, and Lei}]{fb}
Yang Liu, Kun Han, Zhao Tan, and Yun Lei. 2017.
\newblock Using context information for dialog act classification in dnn
  framework.
\newblock In \emph{Proceedings of the 2017 Conference on Empirical Methods in
  Natural Language Processing}, pages 2170--2178. Association for Computational
  Linguistics.

\bibitem[{Luong et~al.(2015)Luong, Pham, and Manning}]{attention}
Minh{-}Thang Luong, Hieu Pham, and Christopher~D. Manning. 2015.
\newblock \href {http://arxiv.org/abs/1508.04025} {Effective approaches to
  attention-based neural machine translation}.
\newblock \emph{CoRR}, abs/1508.04025.

\bibitem[{Mezza et~al.(2018)Mezza, Cervone, Stepanov, Tortoreto, and
  Riccardi}]{trento}
Stefano Mezza, Alessandra Cervone, Evgeny Stepanov, Giuliano Tortoreto, and
  Giuseppe Riccardi. 2018.
\newblock Iso-standard domain-independent dialogue act tagging for
  conversational agents.
\newblock In \emph{Proceedings of the 27th International Conference on
  Computational Linguistics}, pages 3539--3551. Association for Computational
  Linguistics.

\bibitem[{Mikolov et~al.(2018)Mikolov, Grave, Bojanowski, Puhrsch, and
  Joulin}]{fasttext}
Tomas Mikolov, Edouard Grave, Piotr Bojanowski, Christian Puhrsch, and Armand
  Joulin. 2018.
\newblock Advances in pre-training distributed word representations.
\newblock In \emph{Proceedings of the International Conference on Language
  Resources and Evaluation (LREC 2018)}.

\bibitem[{Peitz et~al.(2011)Peitz, Freitag, Mauser, and Ney}]{nmt_punct}
Stephan Peitz, Markus Freitag, Arne Mauser, and Hermann Ney. 2011.
\newblock Modeling punctuation prediction as machine translation.
\newblock In \emph{IWSLT}.

\bibitem[{Rojas{-}Barahona et~al.(2016)Rojas{-}Barahona, Gasic, Mrksic, Su,
  Ultes, Wen, and Young}]{lstmcnn}
Lina~Maria Rojas{-}Barahona, Milica Gasic, Nikola Mrksic, Pei{-}Hao Su, Stefan
  Ultes, Tsung{-}Hsien Wen, and Steve~J. Young. 2016.
\newblock \href {http://arxiv.org/abs/1610.04120} {Exploiting sentence and
  context representations in deep neural models for spoken language
  understanding}.
\newblock \emph{CoRR}, abs/1610.04120.

\bibitem[{Siddhant et~al.(2018)Siddhant, Goyal, and Metallinou}]{transfer}
Aditya Siddhant, Anuj~Kumar Goyal, and Angeliki Metallinou. 2018.
\newblock \href {http://arxiv.org/abs/1811.05370} {Unsupervised transfer
  learning for spoken language understanding in intelligent agents}.
\newblock \emph{CoRR}, abs/1811.05370.

\bibitem[{Stolcke et~al.(2000)Stolcke, Coccaro, Bates, Taylor, Van Ess-Dykema,
  Ries, Shriberg, Jurafsky, Martin, and Meteer}]{stolcke}
Andreas Stolcke, Noah Coccaro, Rebecca Bates, Paul Taylor, Carol Van
  Ess-Dykema, Klaus Ries, Elizabeth Shriberg, Daniel Jurafsky, Rachel Martin,
  and Marie Meteer. 2000.
\newblock \href {https://doi.org/10.1162/089120100561737} {Dialogue act
  modeling for automatic tagging and recognition of conversational speech}.
\newblock \emph{Comput. Linguist.}, 26(3):339--373.

\bibitem[{Thompson et~al.(1993)Thompson, Anderson, Bard, Doherty-Sneddon,
  Newlands, and Sotillo}]{maptask}
Henry~S. Thompson, Anne Anderson, Ellen~Gurman Bard, Gwyneth Doherty-Sneddon,
  Alison Newlands, and Cathy Sotillo. 1993.
\newblock The hcrc map task corpus: Natural dialogue for speech recognition.
\newblock In \emph{Proceedings of the Workshop on Human Language Technology},
  HLT '93, pages 25--30, Stroudsburg, PA, USA. Association for Computational
  Linguistics.

\end{thebibliography}
\bibliographystyle{acl_natbib}

% \appendix

% \section{Supplemental Material}
% \label{sec:supplemental}

% \appendix
\newpage
\clearpage
\section{Appendices}
\subsection{Dialog Act Scheme}
\label{sec:appendix_a}
\begin{tabular}{ |p{3.7cm}|p{5cm}|p{4.8cm}|p{1.5cm}| }
 \hline
 \multicolumn{4}{|c|}{\textbf{Dialog Act - Semantic request}} \\
 \hline
 \textbf{Dialog Act Tag} & \textbf{Description} & \textbf{Example} & \textbf{Count in user utterances (single label only)}\\
  \hline
  \textit{factual question} & factual questions & How old is Tom Cruise; How's the weather today & 360\\
 \hline
 \textit{opinion question} & opinionated questions & What's your favorite book; what do you think of disney movies & 236\\
 \hline
 \textit{yes/no question} & yes or no questions & Do you like pizza; did you watch the game last night & 325\\
 \hline
 \textit{task command} & commands/requests (can be in a question format) for some actions that may be different from the ongoing conversation & can i ask you a question; let's talk about the immigration policy; repeat & 651\\
 \hline
 \textit{invalid command} & general device/system commands that cannot be handled by the social bot & show me a picture; cook food for me & 87\\
 \hline
 \textit{appreciation} & appreciation towards the previous utterance & that's cool; that's really awesome & 201\\
 \hline
 \textit{general opinion} & personal view with polarized sentiment & dogs are adorable; (A: How do you like Tom) B: i think he is great & 2157\\
 \hline
%  \textit{general\_opinion} & personal view with polarized sentiment & (A: do you like animals) B: dogs are adorable; (A: what do you think of ) B: i think he is great; & 2157\\
%  \hline
 \textit{complaint} & complaint about the response from another party & I can't hear you; what are you talking about; you didn't answer my question & 239\\
 \hline
 \textit{comment} & comments on the response from another conversation party & (A: my friend thinks we live in the matrix) B1: she is probably right; B2: you are joking, right; B3: i agree; (A: ... we can learn a lot from movies ...) B: there is a lot to learn; (A: He is the best dancer after michael jackson. What do you think) B: michael jackson & 430\\
 \hline
 \textit{statement non-opinion}   & factual information    & I have a dog named Max; I am 10 years old; (A: what movie have you seen recently) B: the avengers & 1717\\
 \hline
 \textit{other answer} & answers that are neither positive or negative & I don't know;  i don't have a favorite; (A: do you like listening to music) B: occasionally & 428\\
 \hline
 \textit{positive answer} & positive\_answers & yes; sure; i think so; why not & 1278\\
 \hline
 \textit{negative answer} & negative response to a previous question & no; not really; nothing right now & 867\\
 \hline

%  \hline
%  \textit{open\_question} & general question (from SWBD-DAMSL that cannot be mapped to the current scheme) & What's your favorite book\\
 \hline

\end{tabular}
\clearpage
\noindent \begin{tabular}{ |p{3.7cm}|p{5cm}|p{4.8cm}|p{1.5cm}|  }
 \hline
 \multicolumn{4}{|c|}{\textbf{Dialog Act - Functional request}} \\
 \hline
 \textbf{Dialog Act Tag} & \textbf{Description} & \textbf{Example} & \textbf{Count in user utterances (single label only)}\\
 \hline
 \textit{abandon} & not a complete sentence & So uh; I think; can we & 440\\
 \hline
 \textit{nonsense} & utterances that do not make sense to humans & he all out &129\\
 \hline
 \textit{hold} & a pause before saying something & let me see; well & 272\\
 \hline
 \textit{opening} & opening of a conversation & hello my name is tom; hi;  & \\
 \hline
 \textit{closing} & closing of a conversation & nice talking to you; goodbye & 540\\
 \hline
 \textit{thanks} & expression of thankfulness & thank you &80\\
 \hline
 \textit{back-channeling} & acknowledgement to the previous utterance & Uh-huh; (A: i learned that ...) B: okay/yeah/right/really? & 427\\
 \hline
 \textit{apology} & apology & I'm sorry & 29\\
 \hline
 \textit{apology response} & response to apologies & That's all right &6\\
 \hline
 \textit{other} & utterances that cannot be assigned to other tags & & 12 \\
 \hline
\end{tabular}

\clearpage
\subsection{Multi-functionality schemes}
\label{appendix:multi}
\begin{tabular}{ |p{4.3cm}|p{5cm}|p{4cm}|}
 \hline
 \multicolumn{3}{|c|}{\textbf{Multi-label tags}} \\
 \hline
 \textbf{Dialog Act Tags} & \textbf{Example} & \textbf{Count in User Utterances}\\
 \hline
 \textit{positive answer, task command} & (A: wanna know something interesting about it?) B: sure; (A: do you want to talk about some games) B: minecraft & 698\\
 \hline
 \textit{negative answer, task command} & (A: would you like to know more about it) B: I don't want to hear more & 328\\
 \hline
 \textit{task command, general opinion} & (A: what do you want to talk about) B: harry potter stuff & 192 \\
 \hline
  \textit{task command, statement non\_opinion} & let's talk about mario kart & 141\\
 \hline
 \textit{positive answer, statement non\_opinion} & (A: have you read any books recently?) B: I'm reading the great gatsby & 133 \\
 \hline
 \textit{task command, yes/no question} & do you know tom brady; (A: what do you want to talk about?) B: how about movies & 116\\
 \hline
 \textit{negative answer, statement non\_opinion} & (A: do you have pets) B: I don't have any &  66\\
 \hline
 \textit{positive answer, general opinion} & (A: do you like animals) B: My favorite animals is panda & 35 \\
 \hline
 \textit{invalid command, yes/no question} & can you speak louder & 15 \\
 \hline
 \textit{task command, factual question} & what do you know about dodgers &  12 \\
 \hline
 \textit{negative answer, general opinion} & (A: do you watch sports) B: I'm not into sports &  10\\
 \hline
 \textit{task command, opinion question}& (A: what did you find interesting recently) B: what do you think of the new movie & 9 \\
 \hline
 \textit{task command, complaint}& I don't want to hear you talk about anything; would you stop asking me that question & 5 \\
 \hline
 \textit{other answer, general opinion} & (A: what's your favorite movie) B: there are so many to choose from & 5\\
 \hline
 \textit{positive answer, comment} & (A: don't you think so) B: it's true & 4\\
 \hline
 \textit{general opinion, yes/no question} & (A: what would you imagine doing in such situation) B: can we just sleep all day & 3\\
 \hline
 \textit{negative answer, comment} & (A: isn't that interesting) B: that's ridiculous & 3 \\
 \hline
 \textit{general opinion, opinion question} & (A: what book would you recommend me to read) B: how about antifragile & 3\\
 \hline
 
%  \textit{positive\_answer, command, opinion} & (A: would you like to talk about it) B: i like to know  &  \\
%  \hline
%   \textit{device command, yes/no\_question} & &  &  \\
%  \hline
\end{tabular}

\newpage
\clearpage

\subsection{Dialog act tag mapping}\label{mapping}
\begin{tabular}{ |p{3.7cm}|p{5cm}|p{4.8cm}|  }
 \hline
 \textbf{SWBD-DAMSL} & \textbf{SWBD} & \textbf{MIDAS}\\
  \hline
 \textit{statement\_non-opinion} & \textit{sd} & \textit{statement non\_opinion}\\
 \hline
 \textit{Acknowledge (Backchannel)} & \textit{b}  & \textit{back-channeling} \\
 \hline
 \textit{Statement-opinion} & \textit{sv}  & \textit{general opinion} \\
 \hline
 \textit{Agree/Accept} & \textit{aa}  & \textit{pos answer} \\
 \hline
 \textit{Abandoned or Turn-Exit} & \textit{\% -}  & \textit{abandon} \\
 \hline
 \textit{Appreciation} & \textit{ba}  & \textit{appreciation} \\
 \hline
 \textit{Yes-No-Question} & \textit{qy}  & \textit{yes-no question} \\
 \hline
 \textit{Non-verbal} & \textit{x}  & \textit{} \\
 \hline
 \textit{Yes answers} & \textit{ny}  & \textit{pos answer} \\
 \hline
 \textit{Conventional-closing} & \textit{fc}  & \textit{closing} \\
 \hline
 \textit{Uninterpretable} & \textit{\%}  & \textit{abandon} \\
 \hline
 \textit{Wh-Question} & \textit{qw}  &  \\
 \hline
 \textit{No answers} & \textit{nn}  & \textit{neg answer} \\
 \hline
 \textit{Response Acknowledgement} & \textit{bk}  & \textit{back-channeling} \\
 \hline
 \textit{Hedge} & \textit{h}  & \textit{other answers} \\
 \hline
 \textit{Declarative Yes-No-Question} & \textit{qy\^{}d}  & \textit{yes-no question} \\
 \hline
 \textit{Other} & \textit{o,fo,bc,by,fw}  & \textit{other} \\
 \hline
 \textit{Backchannel in question form} & \textit{bh}  & \textit{back-channeling} \\
 \hline
 \textit{Quotation} & \textit{\^q}  & \textit{other opinion} \\
 \hline
 \textit{Summarize/reformulate} & \textit{bf}  & \textit{other opinion} \\
 \hline
 \textit{Affirmative non-yes answers} & \textit{na, ny\^{}e}  & \textit{pos answer} \\
 \hline
 \textit{Action-directive} & \textit{ad}  & \textit{task command} \\
 \hline
 \textit{Collaborative Completion} & \textit{\^{}2}  & \textit{general opinion} \\
 \hline
 \textit{Repeat-phrase} & \textit{b\^{}m}  & \textit{general opinion} \\
 \hline
 \textit{Open-Question} & \textit{qo}  &  \\
 \hline
 \textit{Rhetorical-Questions} & \textit{qh}  &  \\
 \hline
 \textit{Hold before answer/agreement} & \textit{\^{}h}  & \textit{hold} \\
 \hline
 \textit{Reject} & \textit{ar}  & \textit{neg answer} \\
 \hline
 \textit{Negative non-no answers} & \textit{ng,nn\^{}e}  & \textit{neg answer} \\
 \hline
 \textit{Signal-non-understanding} & \textit{br}  & \textit{complaint} \\
 \hline
 \textit{other\_answers} & \textit{no}  & \textit{other answer} \\
 \hline
 \textit{Conventional-opening} & \textit{fp}  & \textit{opening} \\
 \hline
 \textit{Or-Clause} & \textit{qrr}  & \textit{other} \\
 \hline
 \textit{Dispreferred answers} & \textit{arp,nd}  & \textit{neg answer} \\
 \hline
 \textit{3rd-party-talk} & \textit{t3}  & \textit{} \\
 \hline
 \textit{Offers, Options Commits} & \textit{oo,cc,co}  & \textit{other} \\
 \hline
 \textit{Self-talk} & \textit{t1}  & \textit{other} \\
 \hline
 \end{tabular}
 
 \newpage
 \clearpage
 \noindent\begin{tabular}{ |p{3.7cm}|p{5cm}|p{4.8cm}|  }
 \hline
 \textbf{SWBD-DAMSL} & \textbf{SWBD} & \textbf{MIDAS}\\
  \hline
 \textit{Downplayer} & \textit{bd}  & \textit{apology response} \\
 \hline
 \textit{Maybe/Accept-part} & \textit{aap/am}  & \textit{pos answer} \\
 \hline
 \textit{Tag-Question} & \textit{\^{}g}  & \textit{other} \\
 \hline
 \textit{Declarative Wh-Question} & \textit{qw\^{}d}  & \\
 \hline
 \textit{Apology} & \textit{fa}  & \textit{apology} \\
 \hline
 \textit{Thanking} & \textit{ft}  & \textit{thanking} \\
 \hline
 \end{tabular}

\end{document}